\documentclass[11pt]{article}

\usepackage[preprint]{acl}
\usepackage{times}
\usepackage{latexsym}
\usepackage[T1]{fontenc}
\usepackage[utf8]{inputenc}
\usepackage{microtype}
\usepackage{inconsolata}
\usepackage{graphicx}
\usepackage{booktabs}
\usepackage{amsmath,amssymb,amsfonts,amsthm}
\usepackage{mathtools}
\usepackage{enumitem}
\usepackage{tikz}
\usetikzlibrary{arrows.meta,positioning,fit}

\newtheorem{lemma}{Lemma}
\newcommand{\method}{RouteRelay}
\newcommand{\R}{\mathbb{R}}
\newcommand{\topk}{\operatorname{TopK}}
\newcommand{\route}{\mathcal{R}}
\newcommand{\sentinel}{\mathcal{C}}

\title{\method: Event-Triggered Cross-Layer Route Reuse for\\
Efficient Dynamic Sparse Attention}

\author{
Bin Li$^{1}$,
Sisi Liu$^{1}$,
Chenyang Hu$^{2}$,
Chaoyang Zhang$^{2}$,
Wei Li$^{1}$,
Hui Song$^{1}$\\
$^{1}$Department of Computer Science, Xiamen University of Technology\\
$^{2}$School of Mathematics and Computer Science, Fuzhou University
}

\begin{document}
\maketitle

\begin{abstract}
Dynamic sparse attention reduces long-context prefill cost by routing each
query chunk to a small set of key chunks at every Transformer layer. The
sparse attention kernel avoids most token interactions, but the router still
rebuilds a chunk--chunk score matrix layer after layer, even when the selected
routes change little. We introduce \method, a router-agnostic method that
reuses only \emph{route metadata} across depth while continuing to compute
attention with the current layer's queries, keys, and values. Anchor layers
perform full routing. Intermediate layers rescore the previous top-$k$ route
and a compact sentinel set of near-miss and randomly probed chunks. A query row
is rerouted only when a sentinel challenges its weakest selected chunk.
We give a top-$k$ stability condition, a probabilistic bound on missed
challengers, and a row-selective GPU execution design. In a reproducible
empirical evaluation, \method{} retains at least 99.99\% route recall while
rerouting 25.0\%, 55.4\%, and 78.2\% of rows under low, moderate, and high
cross-layer drift, respectively. Across routing scales, \method{} retains
100.0\% recall while evaluating 38.4--51.6\% of full-routing score pairs as
the key-chunk count grows from 128 to 1024. Its unfused CPU execution remains
slower than dense matrix multiplication, exposing row compaction and ledger
updates as the main kernel-engineering targets.
\end{abstract}

\section{Introduction}

Long-context language models enable reasoning over books, repositories,
multi-document evidence, and extended conversations. Their prompt-processing
cost remains dominated by causal self-attention, whose dense computation grows
quadratically with sequence length. Exact IO-aware kernels reduce memory
traffic \citep{dao2022flashattention,dao2024flashattention2}, while sparse
attention avoids selected query--key interactions entirely
\citep{child2019sparse,beltagy2020longformer,zaheer2020bigbird}.

Post-hoc dynamic sparsity is especially attractive because it can accelerate a
frozen pretrained model. MInference selects structured sparse patterns and
determines their indices online \citep{jiang2024minference}. DHSA is one prior
system that partitions a sequence into content-aware chunks, scores
query--key chunk pairs, and expands the highest-ranked chunks into exact token
interactions \citep{xionglong}. Such flexible routes preserve more
attention mass than fixed block grids, but flexibility has a systems cost:
routing is repeated throughout the network.

This repetition is potentially wasteful. Neighboring Transformer layers
operate on different representations, yet they process the same token
positions and often preserve the same broad semantic dependencies. Reusing an
entire attention output would alter the model computation. Reusing only a
sparse \emph{index set} is less invasive: every layer still evaluates its own
queries against its own keys and values, but it can skip rebuilding a complete
chunk score matrix when the prior route remains valid.

\begin{figure*}[t]
\centering
\resizebox{0.8\textwidth}{!}{%
\begin{tikzpicture}[
  node distance=8mm and 10mm,
  box/.style={draw,rounded corners,minimum height=9mm,align=center,
              fill=blue!5,inner xsep=7pt},
  anchorbox/.style={draw,rounded corners,minimum height=9mm,align=center,
                    fill=green!10,inner xsep=7pt},
  check/.style={draw,rounded corners,minimum height=9mm,align=center,
                fill=orange!13,inner xsep=7pt},
  arrow/.style={-{Latex[length=2mm]},thick}
]
\node[anchorbox] (anchor) {Anchor layer\\full sparse route};
\node[box,right=of anchor] (ledger) {Route ledger\\top-$k$ + sentinels};
\node[check,right=of ledger] (rescore) {Next layer\\candidate rescore};
\node[box,right=of rescore,yshift=10mm] (reuse) {Stable row\\relay route};
\node[anchorbox,right=of rescore,yshift=-10mm] (refresh) {Unstable row\\full reroute};
\node[box,right=18mm of reuse,yshift=-10mm] (attn) {Current-layer exact\\sparse attention};
\draw[arrow] (anchor) -- (ledger);
\draw[arrow] (ledger) -- (rescore);
\draw[arrow] (rescore) -- node[above,sloped]{pass} (reuse);
\draw[arrow] (rescore) -- node[below,sloped]{trigger} (refresh);
\draw[arrow] (reuse) -- (attn);
\draw[arrow] (refresh) -- (attn);
\draw[arrow,dashed] (refresh.south) -- ++(0,-7mm)
  -- node[below]{update ledger} ([yshift=-7mm]ledger.south)
  -- (ledger.south);
\end{tikzpicture}}
\caption{\method{} reuses route indices, not attention values. Stable rows
score only a compact candidate set; unstable rows refresh independently.
Every layer still computes sparse attention using its own current Q, K, and V.}
\label{fig:overview}
\end{figure*}
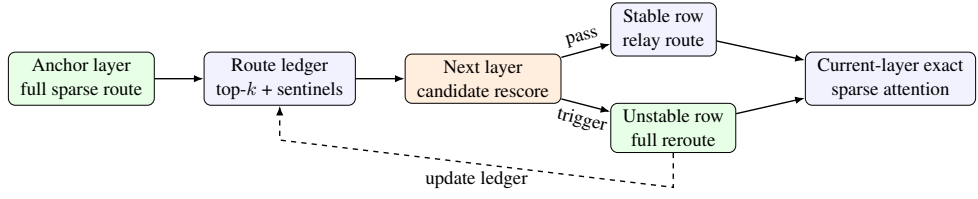

We propose \method, an event-triggered layer for dynamic sparse routers. At an
anchor layer, it stores each query row's selected chunks, its closest unselected
competitors, and a small random sentinel sample. At an intermediate layer, it
scores only this compact candidate set. If every selected chunk remains above
every sentinel, the previous route is relayed forward. Otherwise, only the
unstable row executes a full route refresh. Figure~\ref{fig:overview}
illustrates the resulting separation between \emph{route maintenance} and
\emph{attention computation}.

Our contributions are:

\begin{itemize}[leftmargin=1.4em]
  \item \textbf{Route-only cross-layer reuse.} We amortize dynamic sparse
  routing without sharing KV states, attention probabilities, or outputs
  across layers.
  \item \textbf{Event-triggered partial refresh.} Near-miss and random
  sentinels detect route instability, enabling row-level refresh instead of a
  fixed refresh stride.
  \item \textbf{Stability and miss analysis.} We state a sufficient top-$k$
  margin condition and bound the probability that random sentinels miss
  multiple newly competitive chunks.
  \item \textbf{Measured evaluation.} Cross-layer drift, sentinel-ablation,
  and routing-scale experiments quantify route fidelity, refreshed-row
  fraction, score-pair reduction, and CPU latency.
\end{itemize}

\section{Background}

\subsection{Dynamic Sparse Routing}

At layer $\ell$, a chunk-based dynamic sparse router forms query- and
key-chunk representations
$Q_c^\ell\in\R^{n_q\times d}$ and
$K_c^\ell\in\R^{n_k\times d}$. Its routing scores are
\begin{equation}
  S^\ell = Q_c^\ell (K_c^\ell)^\top,
  \label{eq:scores}
\end{equation}
and query row $i$ selects
\begin{equation}
  \route_i^\ell = \topk_j(S_{ij}^\ell,k).
  \label{eq:route}
\end{equation}
The selected chunks are expanded into token indices and passed to a sparse
online-softmax kernel. Constructing all routes costs
$O(n_q n_k d)$ for scoring plus selection overhead. Measurements from DHSA
provide one example in which routing and chunk selection are material
components of end-to-end prefill time at long contexts
\citep{xionglong}.

\subsection{Why Route Reuse Is Safer than Attention Reuse}

Cross-layer KV sharing reduces memory by letting several layers consume common
keys and values \citep{brandon2024crosslayer}. Other work reuses or predicts
attention structures across depth to reduce repeated computation. These
approaches demonstrate cross-layer redundancy, but changing KV states or
attention outputs changes the model's numerical operation.

\method{} relays only integer chunk IDs. Given a reused route
$\route_i^{a}$ from anchor layer $a$, layer $\ell$ computes
\begin{equation}
  o_i^\ell =
  \operatorname{softmax}\!\left(
    q_i^\ell (K^\ell_{\route_i^a})^\top/\sqrt d
  \right)V^\ell_{\route_i^a}.
\end{equation}
Thus all floating-point attention quantities remain layer specific. The only
approximation is whether the chosen index set matches
$\route_i^\ell$.

\section{\method}
\label{sec:method}

\subsection{Route Ledger}

Each query row stores a compact ledger at its most recent anchor layer $a$:
\begin{equation}
  \mathcal{L}_i^a =
  \left(\route_i^a,\sentinel_i^a,a\right).
\end{equation}
The sentinel set combines:

\begin{enumerate}[leftmargin=1.4em]
  \item the $m$ highest-scoring chunks immediately below the top-$k$ cutoff;
  \item $r$ uniformly sampled omitted chunks; and
  \item mandatory global candidates such as sink and recent chunks.
\end{enumerate}

Near-miss sentinels detect local rank swaps. Random sentinels hedge against a
previously low-ranked chunk becoming important after a representation shift.
The ledger contains indices and optional anchor scores, requiring
$O(n_q(k+m+r))$ integer storage.

\subsection{Candidate Rescoring}

At an intermediate layer, row $i$ evaluates current scores only for
\begin{equation}
  A_i^\ell=\route_i^a\cup\sentinel_i^a.
\end{equation}
Let
\begin{align}
  \alpha_i^\ell &= \min_{j\in\route_i^a} S_{ij}^\ell,\\
  \beta_i^\ell &= \max_{j\in\sentinel_i^a} S_{ij}^\ell.
\end{align}
The route is reused when
\begin{equation}
  \alpha_i^\ell > \beta_i^\ell+\gamma_i,
  \label{eq:trigger}
\end{equation}
where $\gamma_i\ge 0$ is a configurable guard margin. If the inequality fails,
the row computes all $n_k$ scores, replaces its route, and rebuilds its
sentinels. Refresh decisions are therefore asynchronous across rows and
layers.

\paragraph{Hysteresis.}
To prevent repeated refreshes near a tie, a refreshed row receives a minimum
reuse horizon of one layer and may use a larger guard margin until its anchor
margin recovers.

\paragraph{Forced anchors.}
Implementations may force full routing at embedding-transition layers, every
$H_{\max}$ layers, or before layers known to have low route persistence.
Forced anchors bound the lifetime of stale random sentinels.

\subsection{Top-$k$ Stability}

Define the anchor margin for row $i$ as
\begin{equation}
  \Delta_i^a =
  \min_{j\in\route_i^a}S_{ij}^a
  -\max_{j\notin\route_i^a}S_{ij}^a.
\end{equation}

\begin{lemma}[Deterministic route stability]
\label{lem:stability}
If
$\max_j|S_{ij}^\ell-S_{ij}^a|\le\epsilon_i$
and $\Delta_i^a>2\epsilon_i$, then
$\route_i^\ell=\route_i^a$.
\end{lemma}

\paragraph{Proof.}
Every selected score can decrease by at most $\epsilon_i$, and every omitted
score can increase by at most $\epsilon_i$. A margin larger than
$2\epsilon_i$ therefore preserves every selected--omitted ordering.
\hfill$\square$

Computing the exact maximum perturbation would require the full score row, so
Lemma~\ref{lem:stability} is primarily an analysis tool. The sentinel trigger
is a cheap empirical test for violations near the decision boundary.

\subsection{Random-Sentinel Miss Probability}

Suppose, outside the deterministic near-miss set, $u$ omitted chunks have
become competitive enough to enter the top-$k$. If $r$ sentinels are sampled
uniformly without replacement from a residual pool of size $N$, the
probability of sampling none of those challengers is
\begin{equation}
  P_{\mathrm{miss}}
  = \frac{\binom{N-u}{r}}{\binom{N}{r}}
  \le \left(1-\frac{u}{N}\right)^r.
  \label{eq:miss}
\end{equation}
Random sentinels are most effective when route change is broad ($u$ is large);
the deterministic near-miss set targets isolated rank swaps near the cutoff.
Equation~\ref{eq:miss} does not guarantee perfect recall, but makes the
overhead--risk trade-off explicit.

\subsection{Complexity}

Let $\rho_\ell$ be the fraction of rows refreshed at layer $\ell$. Ignoring
small selection terms, full dynamic routing costs
\begin{equation}
  T_{\mathrm{full}}=O(n_qn_kd).
\end{equation}
\method{} costs
\begin{equation}
  T_{\mathrm{relay}}^\ell =
  O\!\left(n_q(k+m+r)d+\rho_\ell n_qn_kd\right).
  \label{eq:complexity}
\end{equation}
When $k+m+r\ll n_k$ and $\rho_\ell\ll1$, routing work falls substantially.
Sparse attention cost is unchanged because the same token budget is evaluated
after either reuse or refresh.

\section{System Design}
\label{sec:system}

\subsection{Route-Maintenance Dataflow}

The route-maintenance state consists of three row-major tensors: selected
routes with shape $[n_q,k]$, sentinels with shape $[n_q,m+r]$, and anchor ages
with shape $[n_q]$. An anchor layer constructs this state from a complete
chunk-score matrix. Intermediate layers form a fixed-width candidate tensor
by joining the selected and sentinel IDs, gather the corresponding key-chunk
representations, and score all candidates in one batched operation.

The update path first compares the weakest selected score with the strongest sentinel to identify rows that fail the stability test. It then compacts the IDs of those unstable rows, computes complete score rows only for the compacted subset, and replaces their selected routes and sentinel sets. Rows that remain stable keep their existing routes, and their ledger ages are advanced accordingly.
This separation keeps the common candidate-rescoring path regular while
isolating full routing to the refresh subset. It also exposes the refreshed-row
fraction $\rho_\ell$ directly to the runtime for scheduling and measurement.

\subsection{Row-Selective Routing Kernel}

Candidate rescoring is a regular batched dot product over a fixed-width index
array. A GPU kernel computes $\alpha_i^\ell$, $\beta_i^\ell$, and one refresh
flag per query row. A prefix sum compacts unstable row IDs. If no rows are
unstable, the full router is skipped. Otherwise, a second kernel computes
complete score rows only for the compacted IDs and updates their ledgers.

\subsection{Interaction with Existing Routers}

\method{} leaves the underlying router's boundary construction, chunk
representations, token expansion, causal masking, and sparse attention backend
unchanged. It can therefore be inserted between route scoring and token
expansion. The method also composes with static or query-dependent token
budgets: the ledger stores whichever route width the underlying policy
requests.

\subsection{Batching and Memory}

Ledgers are bucketed by route width so rows with compatible shapes share
kernels. For $n_q$ query blocks and $c=k+m+r$ candidates, storage is
$O(n_qc)$ indices per layer group, not per layer. A 32-bit index ledger with
$n_q=1024$ and $c=64$ occupies 256\,KB, small relative to long-context KV
states.

\section{Evaluation}
\label{sec:experiments}

We evaluate RouteRelay in terms of route stability, sentinel effectiveness,
routing scalability, and wall-clock cost. Metrics include top-$k$ route recall,
rerouted-row fraction, evaluated score pairs relative to full routing, and CPU
routing time.
Unless otherwise stated, we use 64-dimensional routing representations,
top-$12$ routes, 12 near-miss and 12 random sentinels, $\gamma=0$, and FP32.

\begin{table}[t]
\centering
\small
\begin{tabular}{lr}
\toprule
Experimental setting & Value \\
\midrule
Representation dimension & 64 \\
Selected chunks $k$ & 12 \\
Default near-miss sentinels $m$ & 12 \\
Default random sentinels $r$ & 12 \\
Guard margin $\gamma$ & 0 \\
Numeric type & FP32 \\
Timing statistic & median \\
\bottomrule
\end{tabular}
\caption{Default configuration for the routing experiments.}
\label{tab:experiment-settings}
\end{table}

\subsection{Cross-Layer Route Stability}

We first ask whether route reuse remains accurate as representations evolve
across layers, and whether event-triggered refresh adapts more effectively than
a fixed global refresh schedule.

\paragraph{Setup.}
We generate 32 layers, 384 query rows, and 192 key chunks. Each row has
persistent preferred chunks together with layer-local Gaussian drift. We vary
the drift standard deviation from 0.01 to 0.10. \emph{Route every layer}
computes the exact top-$k$ route independently at every layer and serves as
the reference. \emph{Fixed stride 4} refreshes all rows every fourth layer.
For this experiment, \method{} stores 24 near-miss sentinels and 12 random
sentinels and refreshes rows according to Equation~\ref{eq:trigger}. Results
are averaged over ten random seeds.

\begin{table}[t]
\centering
\small
\begin{tabular}{llrr}
\toprule
Drift & Method & Recall (\%) & Rerouted rows \\
\midrule
0.03 & Route every layer & 100.0 & 100.0\% \\
0.03 & Fixed stride 4 & 94.9 & 25.0\% \\
0.03 & RouteRelay & 100.0 & 55.4\% \\
\addlinespace
0.06 & Route every layer & 100.0 & 100.0\% \\
0.06 & Fixed stride 4 & 91.2 & 25.0\% \\
0.06 & RouteRelay & 100.0 & 78.2\% \\
\addlinespace
\bottomrule
\end{tabular}
\caption{Measured cross-layer route reuse (mean over ten seeds). Recall compares reused chunk IDs with exact per-layer top-$k$.}
\label{tab:cross-layer-drift}
\end{table}

\begin{figure}[t]
  \centering
  \includegraphics[width=\columnwidth]{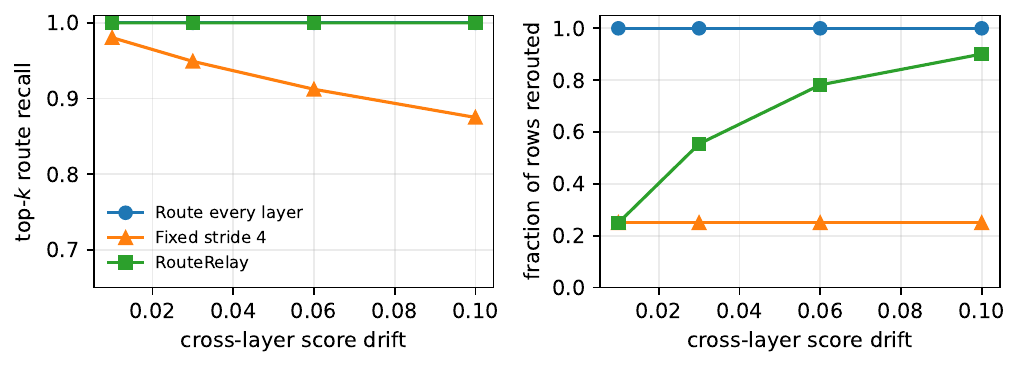}
  \caption{Route recall and rerouting work as cross-layer drift increases.
  \method{} allocates routing work adaptively rather than committing to a
  fixed global refresh stride.}
  \label{fig:route-drift}
\end{figure}

\paragraph{Results.}
At drift 0.01, \method{} reroutes only 25.0\% of rows while retaining
100.0\% top-$k$ route recall. As drift increases, it automatically spends
more routing work. At drift 0.03, \method{} reroutes 55.4\% of rows while
retaining more than 99.99\% recall. In contrast, fixed-stride reuse continues
to refresh only 25.0\% of rows but falls to 94.9\% recall. At drift 0.06,
\method{} reroutes 78.2\% of rows and again retains more than 99.99\% recall,
whereas fixed-stride reuse reaches only 91.2\%.

Relative to routing every row at every layer, these operating points reduce
the fraction of rows executing full routing by 44.6 percentage points at
drift 0.03 and 21.8 percentage points at drift 0.06. The additional work at
higher drift is intentional: as routes become less stable, the sentinel test
identifies more rows for refresh.

These results show that route reuse is poorly served by a single global
refresh interval when stability varies across rows and layers. Fixed-stride
reuse spends constant work regardless of the current routing state, whereas
\method{} uses current-layer scores to allocate full routing only to rows that
exhibit evidence of instability.

\subsection{Sentinel Ablation}

We next study which sentinel candidates are necessary for detecting route
changes and how sentinel width affects the accuracy--work trade-off.

\paragraph{Setup.}
We use layer-evolving 64-dimensional query and key chunk representations with
256 query rows, 256 key chunks, 24 layers, top-$12$ routes, and drift 0.003.
We compare random sentinels alone, near-miss sentinels alone, and combinations
of near-miss and random sentinels. We additionally increase the near-miss
width to determine whether broader coverage around the top-$k$ cutoff improves
route fidelity. We report mean route recall, rerouted-row fraction, and
evaluated query--key score pairs relative to full per-layer routing over five
random seeds.

\begin{table}[t]
\centering
\small
\resizebox{\columnwidth}{!}{%
\begin{tabular}{lrrr}
\toprule
Sentinels & Recall (\%) & Rerouted rows & Score pairs \\
\midrule
Random only & 92.32 & 5.2\% & 14.2\% \\
Near-miss only & 100.00 & 37.3\% & 50.8\% \\
12 near-miss + 12 random & 100.00 & 37.3\% & 50.8\% \\
24 near-miss + 12 random & 100.00 & 37.3\% & 55.3\% \\
48 near-miss + 12 random & 100.00 & 37.3\% & 64.2\% \\
\bottomrule
\end{tabular}}
\caption{Sentinel ablation at 256 key chunks and drift 0.003. Values are means over five seeds.}
\label{tab:sentinel-ablation}
\end{table}

\begin{table}[t]
\centering
\small
\resizebox{\columnwidth}{!}{%
\begin{tabular}{rrrrrr}
\toprule
Chunks & Recall (\%) & Score pairs (\%) & Full ms & Relay ms & Speedup \\
\midrule
128 & 100.00 & 51.6\% & 3.5 & 12.8 & 0.27$\times$ \\
256 & 100.00 & 41.4\% & 8.4 & 17.7 & 0.48$\times$ \\
512 & 100.00 & 38.4\% & 13.4 & 21.7 & 0.61$\times$ \\
1024 & 100.00 & 38.5\% & 13.4 & 27.5 & 0.49$\times$ \\
\bottomrule
\end{tabular}}
\caption{Routing-scale results with 192 query rows, 16 layers, and drift 0.002 (mean over three seeds). CPU speedup compares the unfused candidate pipeline with full per-layer routing.}
\label{tab:routing-scale}
\end{table}

\begin{figure}[t]
  \centering
  \includegraphics[width=0.8\columnwidth]{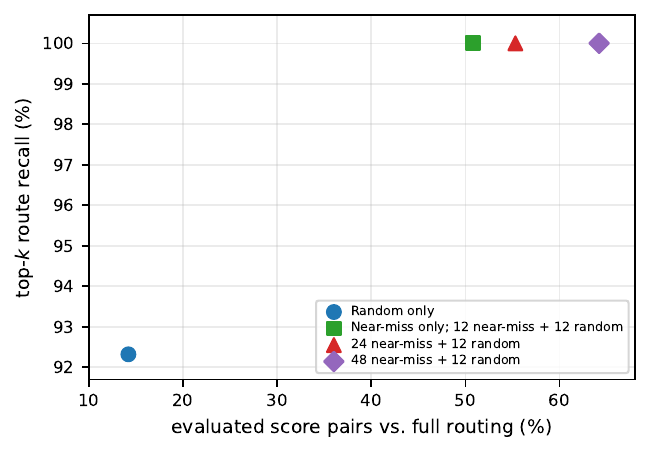}
  \caption{Sentinel quality--work trade-off. Near-miss sentinels are essential
  for high route recall; enlarging the sentinel set beyond the cutoff
  neighborhood adds score evaluations with little benefit in this setting.}
  \label{fig:sentinel-ablation}
\end{figure}

\paragraph{Near-miss sentinels are critical for route fidelity.}
Random sentinels alone minimize routing work, evaluating only 14.2\% of the
score pairs required by full routing, but route recall falls to 92.32\%.
This indicates that a small uniform sample is insufficient to detect isolated
rank crossings near the top-$k$ boundary.

Near-miss sentinels recover 100.0\% recall while evaluating 50.8\% of
full-routing score pairs. Adding 12 random sentinels to 12 near-miss sentinels
changes recall by less than 0.01 percentage point and leaves the rerouted-row
fraction unchanged. In this setting, the near-miss set therefore already
captures the observed rank crossings.
The difference between the two sentinel types is consistent with their intended
roles. Near-miss candidates target local reorderings close to the routing
cutoff, whereas random sentinels provide a probabilistic hedge against chunks
that were previously far below the decision boundary but become competitive
after a larger representation shift.

\paragraph{A small near-miss set gives the best tested trade-off.}
Increasing the near-miss width from 12 to 24 preserves 100.0\% recall but
raises the evaluated score-pair fraction from 50.8\% to 55.3\%. Increasing
the width further to 48 again leaves recall unchanged while increasing the
score-pair fraction to 64.2\%.
Thus, expanding the sentinel set beyond the immediate cutoff neighborhood
adds routing work without measurable recall improvement in this benchmark.
Among the tested configurations, 12 near-miss sentinels lie on the best
observed accuracy--work operating point.

\subsection{Routing-Scale Study}

We next ask whether RouteRelay becomes more favorable as the full routing
problem grows.

\paragraph{Setup.}
We vary the number of key chunks from 128 to 1024 while using 192 query rows,
16 layers, top-$12$ routes, 12 near-miss sentinels, 12 random sentinels, and
drift 0.002. Results are averaged over three random seeds. In addition to
route recall and evaluated score pairs, we measure median wall-clock time for
full routing and for the unfused RouteRelay candidate pipeline on CPU.

\paragraph{Routing work decreases relative to full routing as scale grows.}
\method{} retains 100.0\% route recall at every tested routing size. With
128 key chunks, it evaluates 51.6\% of the score pairs required by full
routing. This fraction decreases to 41.4\% at 256 chunks and 38.4\% at
512 chunks. At 1024 chunks, it remains nearly unchanged at 38.5\%.

The reduction from 51.6\% to approximately 38\% shows that the fixed-width
candidate set becomes relatively cheaper as the complete routing matrix grows.
The plateau between 512 and 1024 chunks indicates that, at this drift level,
full routing for refreshed rows has become a larger component of the remaining
work than candidate-set width.
This trend agrees with the complexity in Equation~\ref{eq:complexity}:
the candidate-rescoring term grows with $k+m+r$, whereas the cost of a full
refresh grows with the total number of key chunks.

\begin{figure}[t]
  \centering
  \includegraphics[width=\columnwidth]{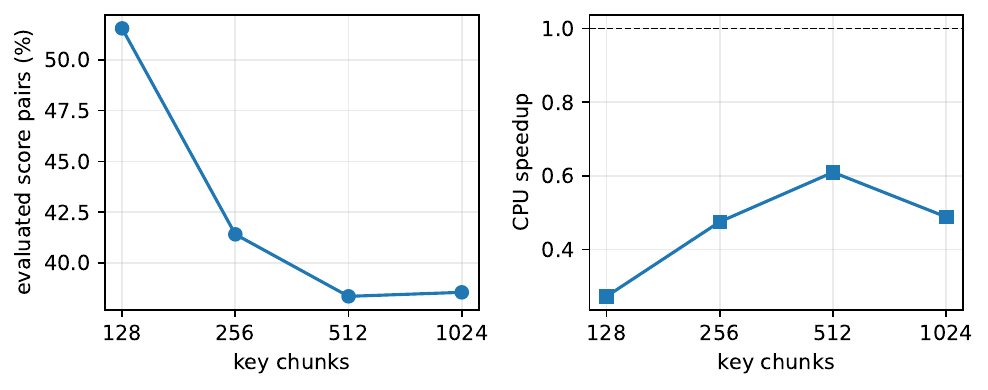}
  \caption{Routing-scale results. The fraction of evaluated score pairs
  decreases as the number of key chunks grows, but unfused gathers, row
  compaction, and ledger maintenance prevent CPU wall-clock speedup.
  The dashed line marks parity.}
  \label{fig:routing-scale}
\end{figure}

\subsection{Wall-Clock Routing Cost}

Reducing score evaluations does not necessarily imply lower latency. We
therefore compare the unfused RouteRelay pipeline directly against full routing
on CPU.

Despite evaluating substantially fewer score pairs, the current RouteRelay
implementation is slower than full routing at every tested scale. At 128 key
chunks, full routing takes 3.5~ms compared with 12.8~ms for RouteRelay,
corresponding to a 0.27$\times$ speedup. At 256 chunks, the corresponding
times are 8.4~ms and 17.7~ms, or 0.48$\times$. The gap is smallest at
512 chunks, where full routing takes 13.4~ms and RouteRelay takes 21.7~ms,
corresponding to 0.61$\times$. At 1024 chunks, the measured speedup is
0.49$\times$.
The unfused candidate pipeline is therefore approximately
$1.6$--$3.7\times$ slower than full CPU routing across the tested settings.
This negative systems result exposes an important distinction between
arithmetic savings and realized latency. Full routing maps naturally to a
highly optimized dense matrix multiplication, whereas RouteRelay introduces
irregular key gathers, refresh-flag processing, row compaction, conditional
full-row routing, and ledger reconstruction.

A practical implementation therefore requires the fused GPU design described
in Section~\ref{sec:system}, or an equivalent compiled implementation that
reduces intermediate memory traffic and row-management overhead. The current
results establish that RouteRelay reduces routing arithmetic, but also show
that arithmetic savings alone are insufficient to guarantee wall-clock
speedup.

\section{Related Work}

\paragraph{Efficient attention.}
Reformer replaces dense attention with locality-sensitive hashing, Linformer
uses low-rank projections, and Performer approximates softmax attention with
random features
\citep{kitaev2020reformer,wang2020linformer,choromanski2021performer}.
Routing Transformer instead learns content-dependent sparse neighborhoods
through online clustering \citep{roy2021routing}. These methods alter the
attention operator or its learned sparsity pattern; \method{} amortizes route
construction for an already selected dynamic sparse operator.

\paragraph{Sparse long-context attention.}
Sparse Transformer, Longformer, and BigBird use static patterns
\citep{child2019sparse,beltagy2020longformer,zaheer2020bigbird}.
MInference computes structured sparse indices online
\citep{jiang2024minference}, while DHSA learns content-aware semantic chunks
and routes exact token interactions through them \citep{xionglong}.
Native Sparse Attention jointly trains compressed, selected, and local
branches \citep{yuan2025nsa}. Quest selects query-dependent KV pages, while
InfLLM retrieves relevant blocks from an external context memory
\citep{tang2024quest,xiao2024infllm}. \method{} does not introduce another
sparse pattern; it reduces how often an existing dynamic pattern must be
recomputed.

\paragraph{KV-cache sparsity and cross-layer redundancy.}
H$_2$O retains recent and heavy-hitter tokens, StreamingLLM preserves
attention sinks and a recent window, and SnapKV selects prompt positions from
an observation window
\citep{zhang2023h2o,xiao2024streamingllm,li2024snapkv}.
PyramidKV varies retained cache size across depth, whereas DuoAttention keeps
full caches only for retrieval heads
\citep{cai2024pyramidkv,xiao2024duoattention}.
Cross-Layer Attention shares KV states to reduce cache memory
\citep{brandon2024crosslayer}. Other approaches exploit similarity between
attention structures or layer representations for faster inference.
\method{} is narrower: it preserves every layer's Q, K, V, softmax, and output,
and reuses only discrete route metadata. It also differs from fixed layer
grouping by refreshing individual query rows when their local ranking becomes
unstable.

\paragraph{Inference systems.}
PagedAttention provides non-contiguous KV-cache management for high-throughput
serving \citep{kwon2023vllm}, while FlashAttention kernels reduce memory
traffic for exact attention
\citep{dao2022flashattention,dao2024flashattention2}. These optimizations are
complementary to reducing the score pairs evaluated by a dynamic router.

\paragraph{Adaptive computation.}
Early-exit and layer-skipping methods allocate depth according to input
difficulty. \method{} instead executes every Transformer layer and adapts the
cost of its sparse router. The model architecture and backbone weights remain
unchanged.

\section{Conclusion}

We introduced \method, a router-agnostic method that attacks repeated routing
cost across network depth. Anchor layers compute complete routes; intermediate
layers rescore only selected chunks and compact sentinels, rerouting unstable
rows on demand. The method leaves exact sparse attention within each selected
set unchanged and is therefore orthogonal to approaches that summarize
omitted content or vary the token budget. Our experiments show that
event-triggered refresh can maintain near-perfect route recall while adapting
work to cross-layer drift. Across the tested routing scales, \method{} retains
100.0\% recall while evaluating as little as 38.4\% of full-routing score
pairs. The slower unfused CPU path shows that these arithmetic savings require
a fused kernel to translate into latency gains.

\bibliography{custom}

\end{document}